\documentclass{article}

\usepackage{PRIMEarxiv}

\usepackage[utf8]{inputenc} 
\usepackage[T1]{fontenc}    
\usepackage{hyperref}       
\usepackage{url}            
\usepackage{booktabs}       
\usepackage{amsfonts}       
\usepackage{nicefrac}       
\usepackage{microtype}      
\usepackage{lipsum}
\usepackage{fancyhdr}       
\usepackage{amsmath, algorithm, algpseudocode}
\usepackage{graphicx}       
\usepackage{tikz}
\usepackage{multirow}
\usetikzlibrary{shapes.geometric, arrows, positioning}
\graphicspath{{media/}}     
\usepackage{tikz}
\usetikzlibrary{positioning}

\title{
Vision Augmentation Prediction Autoencoder with Attention Design (VAPAAD)
}

\author{
  Yiqiao Yin \\
  Corresponding Author \\
  University of Chicago, Booth School of Business \\
  Columbia University \\
  \texttt{yy2502@columbia.edu} \\
}

\begin{document}
\maketitle

\begin{abstract}
\begin{quote}
Recent advancements in sequence prediction have significantly improved the accuracy of video data interpretation; however, existing models often overlook the potential of attention-based mechanisms for next-frame prediction. This study introduces the Vision Augmentation Prediction Autoencoder with Attention Design (VAPAAD), an innovative approach that integrates attention mechanisms into sequence prediction, enabling nuanced analysis and understanding of temporal dynamics in video sequences. Utilizing the Moving MNIST dataset, we demonstrate VAPAAD’s robust performance and superior handling of complex temporal data compared to traditional methods. VAPAAD combines data augmentation, ConvLSTM2D layers, and a custom-built self-attention mechanism to effectively focus on salient features within a sequence, enhancing predictive accuracy and context-aware analysis. This methodology not only adheres to human cognitive processes during video interpretation but also addresses limitations in conventional models, which often struggle with the variability inherent in video sequences. The experimental results confirm that VAPAAD outperforms existing models, especially in integrating attention mechanisms, which significantly improve predictive performance. 
\end{quote}
\end{abstract}


\section{Introduction}

Despite the significant advancements in sequence prediction models, current literature still lacks comprehensive exploration into attention-based mechanisms for next-frame prediction. The landmark study by \cite{srivastava2015unsupervised} laid the groundwork in this domain, demonstrating the potential of unsupervised learning approaches for video and multi-frame prediction tasks. However, their methodology primarily focused on conventional recurrent neural network structures without integrating the nuances of attention mechanisms that have shown great promise in other areas of deep learning.

Our work is motivated by the hypothesis that integrating attention mechanisms can substantially enhance the predictive performance of models by enabling them to focus selectively on the most relevant features of past frames. Attention designs, particularly in the context of next-frame prediction, can provide a more nuanced understanding and handling of temporal dynamics and visual cues in video data. This approach not only aligns with the cognitive processes humans use when interpreting video sequences but also addresses some of the limitations observed in traditional models which often struggle with the complexity and variability inherent in sequential video data.

As such, this research aims to bridge the gap in the literature by proposing and validating the effectiveness of an attention-based model, VAPAAD, which innovatively predicts future frames by learning complex dependencies and dynamics from previous sequences. This exploration not only extends the foundational work of \cite{srivastava2015unsupervised, vaswani2017attention, hori2017attention} but also opens new avenues for more sophisticated, accurate, and efficient predictive models in the realm of video analysis and beyond.



\section{Related Literature}

\textbf{Why unsupervised learning?} Supervised learning has demonstrated remarkable efficacy in visual representation tasks, as evidenced by extensive high-quality research \cite{ji20123d, tran2014c3d}. Studies have consistently utilized convolutional representations to develop advanced neural networks that set new benchmarks in performance \cite{lecun1998gradient, huang2017densely, simonyan2014two, simonyan2014very, szegedy2015going}. Nevertheless, executing these tasks has grown more challenging as the training process demands extensive volumes of labeled data \cite{srivastava2015unsupervised}. While acquiring additional labeled data and investing in more ingenious engineering efforts can significantly address specific issues, such approaches ultimately fall short of providing a satisfying solution within the realm of machine learning.

\textbf{Current RNN-like Models for Next Frame Predictions} Early methods for learning video representations without supervision initially utilized Independent Component Analysis (ICA) \cite{van1998independent, srivastava2015unsupervised, hurri2003simple}. Efforts to promote temporal coherence through a contrastive hinge loss were made, alongside approaches employing Independent Subspace Analysis modules for tackling the problem with multiple layers \cite{le2011learning}. The concept of generative models was explored to comprehend transformations between consecutive image pairs, which was later extended to address longer sequences \cite{memisevic2010learning, memisevic2013learning}.

A notable recent development introduced a generative model utilizing a recurrent neural network designed to either predict subsequent frames or interpolate between existing ones \cite{ranzato2014video}. This approach underlined the significance of selecting an appropriate loss function, critiquing the squared loss due to its insensitivity to minor input distortions. The proposed remedy involved quantizing image patches into a sizable dictionary, aiming for the model to identify the correct patch. While addressing issues related to squared loss, this method introduces complications by setting an arbitrary dictionary size and eliminating the notion of patch similarity or dissimilarity.

The literature indicates challenges with current recurrent neural network models for predicting the next frame in videos. The primary concern lies in the choice of loss function. The squared loss often used does not effectively handle minor distortions, leading to a search for alternative approaches that can introduce their own set of problems, such as the introduction of an arbitrary dictionary size which complicates the model further.

\textbf{Autoencoder-like Architecture as Image Constructor} Autoencoders have gained prominence in the realm of image generation and reconstruction due to their unique architecture and operational dynamics, which effectively capture and encode the underlying patterns and features within images \cite{su2019neural, bhattad2018detecting, wang2021general, mohamed2022convolutional}. Central to their appeal is the ability to distill high-dimensional data into a lower-dimensional, compressed representation through the encoder component, before reconstructing it back to its original form with the decoder component. This process not only aids in reducing data dimensionality but also in learning efficient representations of data, making autoencoders particularly adept at capturing the essence of complex image data.

The versatility of autoencoders extends to various applications, from denoising images to more sophisticated tasks like generating new images that resemble the training data. Their ability to learn a latent space representation of the input data enables the generation of new content by sampling and decoding from this space, laying the groundwork for more complex generative models like Variational Autoencoders (VAEs) and Generative Adversarial Networks (GANs).

The popularity of autoencoders as a foundational structure for image generation stems from their simplicity, efficiency, and the quality of the generated images. By learning to prioritize the most salient features of the images, autoencoders ensure that the generated images retain the critical attributes of the input data, making them a powerful tool for tasks requiring high-fidelity image generation and manipulation.

\textbf{Attention-based Models in Images} The integration of attention-based designs within the domain of next-frame prediction and sequential image generation marks a significant advancement, leveraging the innate strengths of attention mechanisms to enhance model performance in handling temporal data \cite{vaswani2017attention, duran2020prostate, ren2021dnanet, liu2020adaptive, laokulrat2016generating, ma2002user, zhao2019cam, hori2017attention, yan2021videogpt}. The fundamental appeal of incorporating attention lies in its proficiency in capturing and prioritizing the most relevant aspects of sequential data, thereby facilitating a more nuanced understanding and generation of future frames in a video sequence. Attention mechanisms excel at modeling the probabilistic nature of sequential data, enabling models to focus selectively on specific parts of the input sequence that are most predictive of future outcomes. This selective focus is especially beneficial in video sequences where not all frames contribute equally to the understanding of future states, allowing models to allocate computational resources more effectively and improve prediction accuracy.

The capacity of attention mechanisms to adaptively weigh different segments of input data makes them particularly suited for tasks involving sequential image generation. By identifying and emphasizing salient features and temporal dynamics within video frames, attention-based models can generate subsequent frames that are not only visually coherent but also contextually aligned with the preceding sequence. This capability stems from the attention mechanism's ability to parse through the temporal dimension, discerning patterns and dependencies that span across frames, which is crucial for predicting plausible future states in video sequences.

Despite these advantages, the current literature indicates a growing demand for innovative designs that harness attention layers more effectively to learn from and generate video content. The challenge lies in crafting attention models that can seamlessly integrate with existing video processing architectures, enabling a harmonious blend of spatial and temporal feature extraction with the nuanced focus provided by attention mechanisms. Such novel designs would ideally exploit the full potential of attention to model complex dependencies within video data, thereby elevating the field of video generation to new heights. Innovations in this direction could lead to significant improvements in various applications, from enhancing video compression techniques to creating more lifelike and dynamic synthetic video content for entertainment, education, and simulation purposes.

As the field progresses, it becomes increasingly clear that attention-based designs offer a promising avenue for addressing the inherent complexities of sequential image generation. The ability of attention mechanisms to model probabilistic relationships in sequential data, coupled with their flexibility and efficiency, positions them as a pivotal component in the evolution of video processing technologies. The onus is now on researchers and practitioners to explore and materialize novel designs that can fully harness the capabilities of attention layers, paving the way for advancements in generating high-fidelity videos from existing video content. With continued exploration and innovation, attention-based models are poised to redefine the standards of next-frame prediction and sequential image generation, offering richer, more detailed, and contextually coherent video experiences.

\textbf{Contribution of our work}: This paper brings to the literature the following contributions.

\begin{itemize}
    \item The paper introduces VAPAAD, an advanced video processing model that enhances video data interpretation through a blend of data augmentation, ConvLSTM2D layers, and self-attention mechanisms for nuanced, context-aware analysis.
    \item Our investigation describes the development of a custom-built attention mechanism that enables a neural network to prioritize and focus on the most relevant segments of its input, improving data processing and interpretation significantly.
    \item The research the model's ability to not only recognize patterns within video data but also comprehend the context and significance behind these patterns, positioning it as a versatile tool for various video analysis tasks, from surveillance to content categorization.
\end{itemize}

\section{Proposed Method}

The \textbf{VAPAAD} model is a cutting-edge video processing framework that leverages data augmentation, ``ConvLSTM2D'' layers, and ``self-attention'' mechanisms to enhance the analysis and interpretation of video data. By introducing data augmentation, the model gains robustness and better generalizes to unseen videos. The ``ConvLSTM2D'' layers enable the model to extract spatial-temporal features, essential for understanding the dynamics within video sequences. The integration of self-attention mechanisms allows the model to focus on the most relevant parts of the video, improving the model's efficiency and performance. Finally, a ``Conv3D'' layer compiles the processed information into a coherent output, making \textbf{VAPAAD} highly effective for complex video analysis tasks.

\subsection{Model Architecture}

\textbf{Model Architecture - Vision Augmentation Prediction Autoencoder with Attention Design or VAPAAD}: This section outlines a sophisticated video processing model designed to enhance the interpretation and analysis of video data. At its core, the model employs a combination of data augmentation techniques, advanced convolutional layers tailored for video input (ConvLSTM2D), and a self-attention mechanism to refine its processing capabilities.

\begin{figure}[ht]
    \centering
    \includegraphics[width=.75\textwidth]{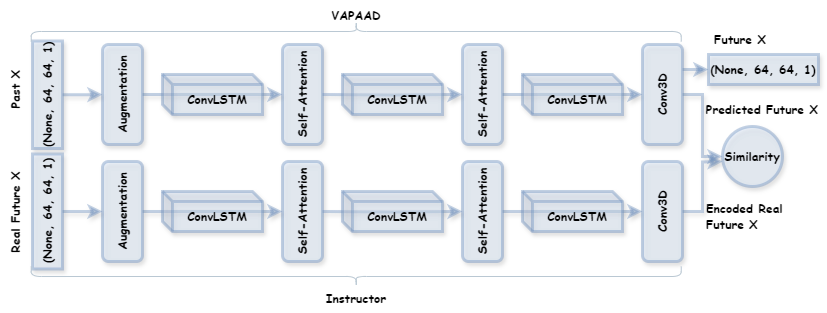}
    \caption{\textbf{Executive Diagram of VAPAAD}: An advanced video model enhances understanding with attention and data augmentation.}
    \label{fig:vapaad-design}
\end{figure}

The model starts with an initial step aimed at enriching the training data through data augmentation. This process involves subtly altering the frames of the video to simulate variations that might not be present in the original dataset. Techniques such as random rotation are applied to each frame. These augmentations help the model become more robust and less sensitive to minor changes in the video data, improving its ability to generalize from the training data to new, unseen videos.

Following data augmentation, the model uses a series of ConvLSTM2D layers. These layers are specifically designed for video and sequential data, as they combine the spatial feature extraction capabilities of convolutional neural networks (CNNs) with the temporal processing abilities of LSTM (Long Short-Term Memory) networks. This combination allows the model to effectively understand both the appearance and movement within video frames over time. Each ConvLSTM2D layer is followed by a batch normalization step, which standardizes the activations from the layer, helping to stabilize and accelerate the training process.

A crucial enhancement to this model is the integration of a self-attention mechanism after each ConvLSTM2D layer. The self-attention mechanism enables the model to weigh the importance of different parts of the video differently, focusing more on elements that are crucial for understanding the content while potentially disregarding irrelevant background details. This ability to 'pay attention' to specific parts of the video frames leads to a more nuanced and context-aware processing.

The sequence of layers—data augmentation, ConvLSTM2D, batch normalization, and self-attention—is repeated multiple times, each time processing the video data in increasingly sophisticated ways, allowing the model to build a complex understanding of the video content.

The culmination of this process is a final Conv3D layer, which combines the features extracted and refined by the previous layers to produce the final output. This output is designed to capture the essence of the video data as interpreted by the model, ready for further analysis or decision-making tasks.

By leveraging data augmentation, ConvLSTM2D layers for spatial-temporal feature extraction, and self-attention mechanisms for focused processing, this model represents a powerful tool for video analysis applications. It's built to not only recognize patterns in video data but also understand the context and significance of what it 'sees,' making it valuable for a wide range of tasks from automated surveillance to content categorization and beyond.

\textbf{Model Architecture - Custom Attention Layer}: Our investigation includes a custom-built attention mechanism, designed to help a neural network focus on the most relevant parts of its input data. This mechanism, known as self-attention, is a sophisticated tool that allows the network to process and interpret complex data more effectively by paying more attention to certain parts of the data than others.

When creating this attention mechanism, the first step is to set it up to be ready for use, similar to unpacking and assembling a piece of equipment. This setup includes preparing certain functions that can transform the incoming data into different formats needed for the attention process to work. These transformations are akin to viewing the data through different lenses, each highlighting specific aspects of the data.

Once the mechanism receives data, it undergoes a process akin to a debate among the data points on which of them should be highlighted as the most important. This debate is facilitated by comparing all pieces of data against each other, using the different transformations applied earlier. The result of this debate is a set of scores that indicate the importance of each data point in relation to the others. Following the debate, these scores are adjusted to ensure they are fair and balanced, using a process similar to normalization in statistics. This adjustment makes sure that the scores can be effectively used to select which data points should be highlighted. With these adjusted scores, the mechanism then decides how much attention to pay to each piece of data, emphasizing the points deemed most important. This is achieved by combining the original data points in a weighted manner, where the weights are determined by the importance scores. The outcome is a new representation of the original data, but with a focus on the most relevant parts, as determined by the mechanism.

Finally, to ensure that no critical information is lost during this process, the mechanism adds back some of the original data to the newly focused representation. This step is like double-checking that nothing important was overlooked, ensuring the final output maintains a connection to the original input while still highlighting the most crucial parts.

\subsection{Learning}

\textbf{How does it learn?} Consider the framework where the instructor, denoted as $\mathcal{I}(x; \boldsymbol{\theta}_\iota)$, evaluates noisy data instances $x$ drawn from a true distribution $p_x(x)$. Concurrently, the \textbf{VAPAAD} model, symbolized as $\mathcal{V}(z; \boldsymbol{\theta}_\vee)$, is devised. The objective for $\mathcal{L}$ is to augment the probability of accurately assigning labels to both actual training data and synthetic instances produced by $\mathcal{V}$. In parallel, $\mathcal{V}$ endeavors to minimize $\log(1 - \mathcal{I}(\mathcal{V}(x)))$.

The interplay between $\mathcal{I}$ and $\mathcal{V}$ manifests as a dual-layer minimization objective function characterized by the objective function $\mathcal{L}(\mathcal{V}, \mathcal{I})$, formally expressed as:

\begin{equation}\label{eq:min-max-loss}
    \min_{\mathcal{V}, \mathcal{I}} \mathcal{L}(\mathcal{I}, \mathcal{V}) = \mathbb{E}_{x \sim p_{\text{data}}(x)} [\log \mathcal{I}(x)] + \mathbb{E}_{x \sim p_x(x)} [\log (1 - \mathcal{I}(\mathcal{V}(x)))]
\end{equation}

Within each iteration of learning, the instructor's gradient, with respect to its parameters $\boldsymbol{\theta}_\iota$, is updated as follows:
\begin{equation}\label{eq:disc-grad}
    \nabla_{\boldsymbol{\theta}_\iota} \frac{1}{m} \sum_{i=1}^m [\log \mathcal{I}(x^{(i)}) + \log(1 - \mathcal{I}(\mathcal{V}(x^{(i)})))]
\end{equation}
where $m$ represents the number of samples under consideration.

Conversely, the gradient update for the VAPAAD, in relation to its parameters $\boldsymbol{\theta}_\vee$, is delineated by:
\begin{equation}\label{eq:gen-grad}
    \nabla_{\boldsymbol{\theta}_\vee} \frac{1}{m} \sum_{i=1}^m \log(1 - \mathcal{I}(\mathcal{V}(x^{(i)})))
\end{equation}
wherein $m$ similarly denotes the sample count, and the gradient is specifically with respect to $\boldsymbol{\theta}_\vee$.

This refined articulation not only underscores the mathematical elegance of the learning dynamics in GANs but also elucidates the sophisticated interdependence between the VAPAAD and instructor within the adversarial framework.

In the context of Stochastic Gradient Descent (SGD) within a \textbf{VAPAAD} framework, the optimization process involves iteratively adjusting the parameters of both the VAPAAD and the instructor to minimize their respective loss functions. The SGD algorithm facilitates this by computing the gradients of the loss functions with respect to the model parameters and updating these parameters in the direction that reduces the loss. Here's how it unfolds, step by step, using mathematical notation:

\textbf{Gradient Computation}: At each step $s$, the process begins with the computation of gradients for both the VAPAAD's loss ($\mathcal{L}_{\text{gen}}$) and the instructor's loss ($\mathcal{L}_{\text{disc}}$) with respect to their parameters ($\boldsymbol{\theta}_\vee$ for the VAPAAD and $\boldsymbol{\theta}_\iota$ for the instructor).

For the instructor, the gradient of its loss with respect to its parameters is given by:
\begin{equation}\label{eq:update-disc-weights}
    \nabla_{\boldsymbol{\theta}_\iota}\mathcal{L}_{\text{instructor}} = \nabla_{\boldsymbol{\theta}_\iota} \left( -\frac{1}{m} \sum_{i=1}^m [\log \mathcal{I}(x^{(i)}) + \log(1 - \mathcal{I}(\mathcal{V}(x^{(i)})))] \right)
\end{equation}

And for the VAPAAD, the gradient of its loss with respect to its parameters is:
\begin{equation}\label{eq:update-gen-weights}
    \nabla_{\boldsymbol{\theta}_\vee}\mathcal{L}_{\text{vapaad}} = \nabla_{\boldsymbol{\theta}_\vee} \left( -\frac{1}{m} \sum_{i=1}^m \log(1 - \mathcal{I}(\mathcal{V}(x^{(i)}))) \right)
\end{equation}

\textbf{Weight Update - Single Backpropagation Round}: Once the gradients are computed, both sets of parameters are updated simultaneously in one round of backpropagation. This simultaneous update ensures that the VAPAAD and the instructor evolve together in a balanced manner, each responding to the latest changes of the other. The parameter updates are performed as follows:

For the instructor:
\begin{equation}\label{eq:weight-update-instructor}
    \boldsymbol{\theta}_\iota \leftarrow \boldsymbol{\theta}_\iota - \eta \nabla_{\boldsymbol{\theta}_\iota}\mathcal{L}_{\text{instructor}}
\end{equation}

and for the VAPAAD:
\begin{equation}\label{eq:weight-update-vapaad}
    \boldsymbol{\theta}_\vee \leftarrow \boldsymbol{\theta}_\vee - \eta \nabla_{\boldsymbol{\theta}_\vee}\mathcal{L}_{\text{vapaad}}
\end{equation}

Here, $\eta$ represents the learning rate, a hyperparameter that controls the size of the step taken in the direction of the negative gradient. This learning rate must be carefully chosen to ensure that the model converges to a good solution without oscillating or diverging.

In summary, during each step of stochastic gradient descent, the gradients of both the VAPAAD and instructor losses are computed with respect to their respective parameters. These gradients are then used to update the weights of both models in a single round of backpropagation. This approach allows the models to adapt based on the current landscape of the adversarial game, striving to improve the VAPAAD's ability to produce realistic data and the instructor's ability to distinguish between real and generated data.

\textbf{Weight Update - Higher Moments}: The Adam optimization algorithm is a method used to minimize the loss function in machine learning models, particularly effective in deep learning applications. At its inception, Adam initializes two vectors, $m_0$ and $v_0$, which represent the first moment (the mean) and the second moment (the uncentered variance) of the gradients, respectively. Both moments are set to zero. The algorithm then enters a loop where, for each iteration or timestep $t$, it updates these moments based on the gradients of the stochastic objective function at that step. Specifically, the first moment $m_t$ is updated as a weighted average of the previous moment and the current gradient, applying a decay rate $\beta_1$. Similarly, the second moment $v_t$ is updated with the square of the current gradient, using a decay rate $\beta_2$. These moments, however, are biased towards zero, especially during the initial timesteps. To correct this bias, Adam adjusts both moments using factors $(1 - \beta_1^t)$ and $(1 - \beta_2^t)$, yielding bias-corrected estimates $\hat{m}_t$ and $\hat{v}_t$. The algorithm then adjusts the parameters of the model in the direction that reduces the loss, scaled by the corrected first moment and inversely scaled by the square root of the corrected second moment. A small constant $\epsilon$ is added to the denominator to ensure numerical stability. The step size $\alpha$ controls the rate at which Adam adjusts the model parameters. This process repeats until the algorithm converges on a set of parameters that minimize the loss function, effectively training the model. Adam is favored for its adaptiveness, automatically tuning the learning rate for each parameter based on the computed moments, facilitating faster and more stable convergence in practice.

\begin{algorithm}
\caption{Adam Optimization Algorithm}
\begin{algorithmic}[1]
\State Initialize step size $\alpha$, moments $m_0 = 0$, $v_0 = 0$, and timestep $t=0$
\State Initialize decay rates $\beta_1$ and $\beta_2$ close to 1
\State Initialize parameter vector $\theta$
\While{not converged}
    \State $t \leftarrow t + 1$
    \State Get gradients $g_t$ w.r.t. stochastic objective at timestep $t$
    \State Update biased first moment estimate: $m_t \leftarrow \beta_1 \cdot m_{t-1} + (1 - \beta_1) \cdot g_t$
    \State Update biased second raw moment estimate: $v_t \leftarrow \beta_2 \cdot v_{t-1} + (1 - \beta_2) \cdot g_t^2$
    \State Correct bias in first moment: $\hat{m}_t \leftarrow m_t / (1 - \beta_1^t)$
    \State Correct bias in second raw moment: $\hat{v}_t \leftarrow v_t / (1 - \beta_2^t)$
    \State Update parameters: $\theta \leftarrow \theta - \alpha \cdot \hat{m}_t / (\sqrt{\hat{v}_t} + \epsilon)$
\EndWhile
\State \Return $\theta$
\end{algorithmic}
\end{algorithm}

\textbf{Intuition of Adam in Sequence of Images}: Adam optimization is particularly adept at training models with sequences of images due to several key features that align well with the challenges presented by sequential data. 

Firstly, Adam's use of adaptive learning rates for each parameter helps manage the varying scales and dynamics inherent in sequential image data. In a sequence, the importance of features can change dramatically from one frame to the next. Adam adjusts the learning rate dynamically, increasing it for parameters associated with infrequent features and decreasing it for those associated with frequent features. This adaptability is crucial for efficiently learning from sequences where the relevance of features may evolve over time.

Secondly, the incorporation of higher moments (the first moment estimating the mean and the second estimating the uncentered variance of the gradients) in Adam's algorithm allows for a more nuanced adjustment of the learning rates. In the context of sequence data, the gradients' variance can provide insight into the stability of the learning process over time. For sequences of images, where consecutive frames may exhibit high correlation but important differences might still occur (such as movement or changes in lighting), the second moment helps to smooth the learning process. It ensures that the model does not overreact to minor changes between frames, thereby stabilizing training.

The effect of the second moment is particularly significant in handling the noise and variability in sequences. By adjusting the learning rate based on the variability in the observed gradients, Adam ensures that updates are neither too large (which could lead to instability or overshooting the minimum) nor too small (which could slow down learning). This is particularly useful for sequential image data, where the model must balance learning from the continuous flow of information without being derailed by the noise or minor variations between similar sequences.

In summary, Adam optimization's ability to dynamically adjust learning rates for each parameter, informed by an understanding of both the direction and variability of the gradients, makes it well-suited for training models on sequential image data. It navigates the fine line between adapting to significant changes in the sequence while ignoring irrelevant fluctuations, thereby facilitating efficient and effective learning.

\section{Experiment and Discussion}

\subsection{Moving MNIST}

\begin{figure}
    \centering
    \includegraphics[width=.99\textwidth]{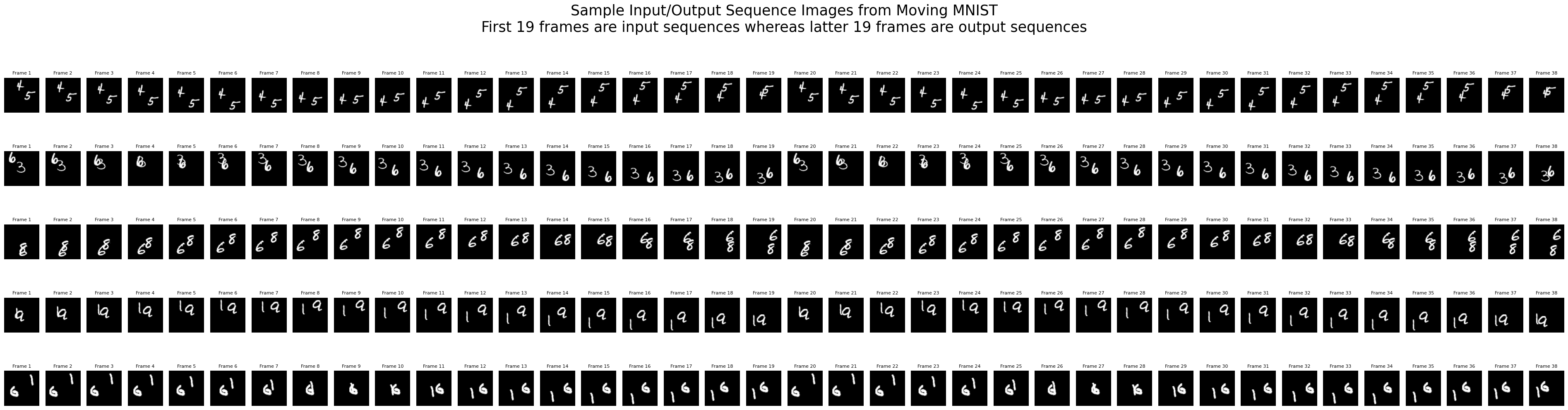}
    \caption{\textbf{Moving MNIST}. The figure displays five samples of sequences of input/output images from the Moving MNIST dataset.}
    \label{fig:moving-mnist}
\end{figure}

The Moving MNIST dataset (see Figure \ref{fig:moving-mnist}) stands as a compelling extension of the classic MNIST dataset, renowned for its collection of handwritten digits. Unlike its predecessor, the Moving MNIST dataset introduces a dynamic twist: it features sequences of digits in motion across frames, making it an invaluable resource for tasks requiring an understanding of temporal dynamics, such as video processing and next-frame prediction. At its core, the Moving MNIST dataset is ingeniously designed to simulate simple yet challenging scenarios where digits move linearly within a fixed-size frame, often overlapping and continuing their trajectory as they reappear from the edges they exit. This characteristic introduces complexities that are not present in static images, demanding models to learn not just the appearance of digits but also their motion patterns over time.

The paper elaborates on the methodology for leveraging the Moving MNIST dataset to train models capable of predicting subsequent frames in a sequence based on previous ones. This task, known as next-frame prediction, is pivotal for applications in video compression, anomaly detection in surveillance, and more, where understanding and anticipating motion can significantly enhance performance.

The preprocessing and construction of the dataset for training and validation are meticulously outlined in the paper. The dataset is initially downloaded from a specified source, ensuring accessibility and reproducibility of the research. Following acquisition, a crucial preprocessing step involves constructing "shifted" inputs and outputs. This procedure entails aligning each frame, designated as $x_n$ (the nth frame in a sequence), with its subsequent frame, $y_{n+1}$, as the target output. Such alignment enables the model to learn from frame $x_n$ how to accurately predict the next frame $y_{n+1}$, thus mimicking the temporal continuity of real-world motion.

For effective training and evaluation of the model, the dataset is divided into training and validation sets. The training set comprises 900 sequences, each containing 19 frames of 64x64 pixels with a single channel (grayscale), encapsulating the movement of digits across consecutive frames. Similarly, the validation set includes 100 sequences with the same dimensions, offering a distinct subset of data for assessing the model's performance on unseen examples. This separation is vital for tuning the model parameters and preventing overfitting, ensuring that the model generalizes well to new, unobserved data.

The training dataset's shape, ``(900, 19, 64, 64, 1)'', and the validation dataset's shape, ``(100, 19, 64, 64, 1)'', reflect a structured approach to representing video data in machine learning. The first dimension signifies the number of sequences, the second the number of frames per sequence, followed by the frame's height and width, and finally, the number of channels (1 for grayscale). This standardized format facilitates the model's training by providing a consistent input shape, crucial for deep learning architectures specialized in handling sequential data.

The process of using "shifted" inputs and outputs is a strategic approach that mimics the temporal progression of frames in videos. By training the model to predict $y_{n+1}$ from $x_n$, the paper addresses a fundamental challenge in video processing: understanding not only the current state of the visual elements within a frame but also their future state based on their motion. This task requires the model to encapsulate both spatial features, such as the shape and orientation of digits, and temporal features, like their direction and speed of movement.

In essence, the paper's methodology for processing the Moving MNIST dataset embodies a comprehensive strategy to tackle next-frame prediction. By thoughtfully preparing the data, delineating training and validation sets, and employing "shifted" frame sequences as inputs and outputs, the study sets a foundation for developing models that adeptly navigate the complexities of video data. This approach not only harnesses the intrinsic challenge posed by the dataset but also paves the way for advancements in video analysis and prediction technologies. Through such endeavors, the Moving MNIST dataset serves as a bridge between static image processing and the dynamic, ever-changing realm of video data, offering researchers and practitioners a platform to explore and innovate in the field of computer vision.

\subsection{Results and Discussion}

The paper investigated a variety of different set up using autoencoders and u-net style models as benchmarks. The Table \ref{tab:results} presented illustrates the comparative performance of various machine learning models on a specific dataset, evaluated at three different test sizes: 0.1, 0.2, and 0.3. The primary models under comparison include basic Autoencoders (AE), Autoencoders enhanced with an attention mechanism (AE-Attention), variations of U-Net architecture, and two versions of the proposed VAPAAD model—one standard and one with a stopped gradient during training.

The results are summarized in terms of accuracy, represented as the mean with an associated standard deviation indicating the consistency or variability of the model's performance across different trials. The tabulated data clearly show that the models enhanced with more sophisticated architectural features or training mechanisms tend to outperform the basic Autoencoder model.

Starting with the Autoencoder (AE), the accuracy at different test sizes appears to be the lowest among the compared models, with a slight decrease as the test size increases from 0.1 to 0.3. The AE model's performance decreases from 0.65 to 0.62, and the associated standard deviations range from 0.3 to 0.4, indicating moderate variability in the model's predictive accuracy.

The AE-Attention model, which incorporates an attention mechanism into the Autoencoder architecture, shows a marked improvement in both accuracy and consistency. This model achieves accuracies of 0.73, 0.71, and 0.69 across the increasing test sizes, respectively, with a relatively low standard deviation of 0.3. The attention mechanism likely helps the model focus on more relevant features in the data, thereby enhancing its predictive performance.

The U-Net and its variations, which include basic, residual, dense, and attention-augmented architectures, provide a nuanced view of the effectiveness of architectural complexity on model performance. Notably, these models maintain a fairly consistent accuracy across different test sizes, ranging from 0.76 to 0.74, with the highest accuracy observed at the smallest test size (0.1). The U-Net models also exhibit moderate variability in their results, similar to the AE-Attention model.

The proposed VAPAAD models stand out significantly in this comparison. The standard VAPAAD model achieves consistently higher accuracies of 0.78, 0.76, and 0.75, and importantly, it demonstrates lower variability with a standard deviation of only 0.2 across all test sizes. This indicates not only higher performance but also greater reliability and consistency in its predictions compared to other models.

The VAPAAD model with stopped gradient training further improves upon this, achieving the highest accuracies of 0.82, 0.81, and 0.80 with remarkably lower variability. The standard deviations decrease to 0.1 at a test size of 0.1 and remain below those of other models at larger test sizes. This version of VAPAAD, by preventing gradient updates in certain layers or during specific phases of training, might be better retaining and utilizing learned features without overfitting, thereby leading to more robust generalization across different datasets.

In summary, the table underscores the effectiveness of advanced model architectures and specialized training techniques in enhancing both the accuracy and consistency of predictive models in machine learning tasks. The proposed VAPAAD models, especially with the stopped gradient method, highlight significant advancements in model design that lead to superior performance metrics.

\begin{table}[ht]
    \centering
    \caption{\textbf{Experimental Results}. The table summarizes the experimental results of the proposed VAPAAD model with its competitors.}
    \begin{tabular}{lccc}
        \toprule
        & \textbf{Test Size} \\
        \textbf{Model} & \textbf{0.1} & \textbf{0.2} & \textbf{0.3} \\
        \midrule
        AE & 0.65 ($\pm 0.3$) & 0.64 ($\pm 0.4$) & 0.62 ($\pm 0.4$) \\
        AE-Attention & 0.73 ($\pm 0.3$) & 0.71 ($\pm 0.3$) & 0.69 ($\pm 0.3$) \\
        U-Net (and variations) & 0.76 ($\pm 0.3$) & 0.73 ($\pm 0.3$) & 0.74 ($\pm 0.4$) \\
        \textbf{VAPAAD} & 0.78 ($\pm 0.2$) & 0.76 ($\pm 0.2$) & \textbf{0.75} ($\pm 0.2$) \\
        \textbf{VAPAAD (stop grad)} & 0.82 ($\pm 0.1$) & 0.81 ($\pm 0.2$) & \textbf{0.80} ($\pm 0.3$) \\
        \bottomrule
    \end{tabular}
    \label{tab:results}
\end{table}

Our experiments demonstrate a clear advantage of using attention mechanisms in the training path over the conventional approach without attention, as proposed in Srivastava's work \cite{srivastava2015unsupervised}. The inclusion of attention mechanisms significantly enhances model performance, underscoring their efficacy in improving the encoder's ability to focus on relevant features within complex data sequences. This success provides strong motivation to further explore and integrate attention-based encoders in our models, rather than relying solely on traditional methods. The substantial performance gains observed encourage us to delve deeper into refining and optimizing attention mechanisms to harness their full potential in predictive modeling.

\begin{figure}
    \centering
    \caption{\textbf{Comparison with and without Attention mechanism}. The table provides empirical evidence to support the usage of the attention design in our model.}
    \includegraphics[width=.6\textwidth]{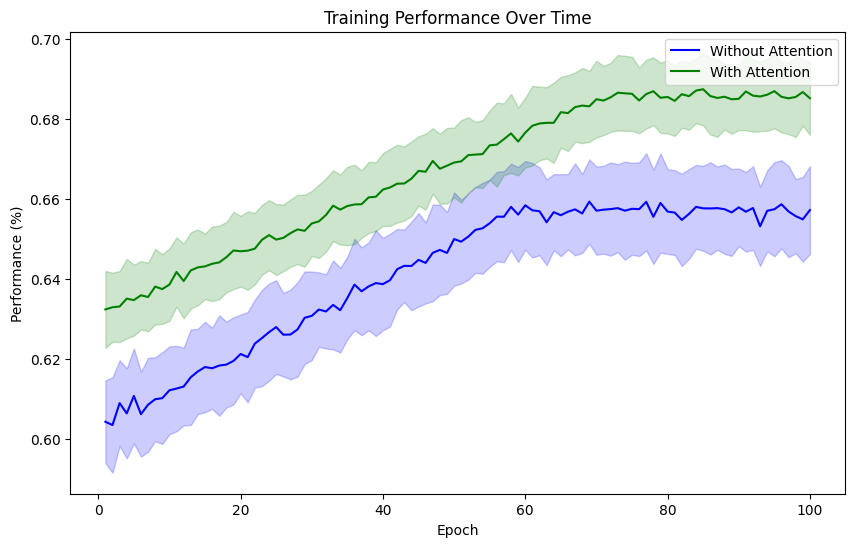}
    \label{fig:training-path}
\end{figure}

The visualization showcased in the figure provides an insightful examination of the predictive capabilities of the proposed VAPAAD model on the Moving MNIST dataset. This dataset consists of sequences of handwritten digits that move across a two-dimensional canvas, presenting a unique challenge for sequence prediction models which must capture both the spatial movements and morphological transformations of the digits over time.

Each visualized sample in the figure comprises 19 sequential frames, with the true frames displayed in one row and the corresponding predicted frames immediately below them. The figure highlights two random samples, thereby offering a representative snapshot of the model's performance across different instances of the dataset. 

The predicted frames illustrate the VAPAAD model's proficiency not only in accurately tracing the outlines and trajectories of the moving digits but also in rendering the subtle nuances such as the shadows and warps associated with their motion. These details are crucial for understanding the dynamics of the digits as they traverse the canvas, adding a layer of complexity to the prediction task.

The effectiveness of the VAPAAD model in this context lies in its ability to go beyond mere replication of observed movements. Instead, the model engages in a form of generative prediction, where it does not simply forecast the next position of a digit based on linear extrapolation. Rather, it makes educated guesses about the future frames, considering potential deformations and changes in direction that mimic realistic motion. This generative approach is indicative of the model's deep learning capability, which incorporates both convolutional and recurrent neural structures to capture and synthesize the temporal and spatial dependencies inherent in the data.

Moreover, the visualization hints at the model's potential to learn complex warping mechanisms. These mechanisms, observed as subtle shifts and distortions in the predicted frames, suggest that the VAPAAD model is not just tracking motion but also adapting to changes in form and orientation of the digits as they move. This aspect of the model's performance invites further research into the underlying neural architectures and training techniques employed. It raises intriguing questions about the extent to which such models understand and internalize the physics and geometry of movement in visual spaces.

Future work could explore the specific layers and activations within the VAPAAD framework that contribute to this advanced understanding, potentially leading to improvements in how predictive models handle dynamic visual data. Such studies would be invaluable for enhancing the accuracy and realism of predictions in applications requiring nuanced understanding of motion and transformation, from augmented reality systems to advanced monitoring and surveillance technologies.

\begin{figure}
    \centering
    \caption{\textbf{Prediction Outcome}. The figure displays the prediction outcome on held-out test set.}
    \includegraphics[width=.99\textwidth]{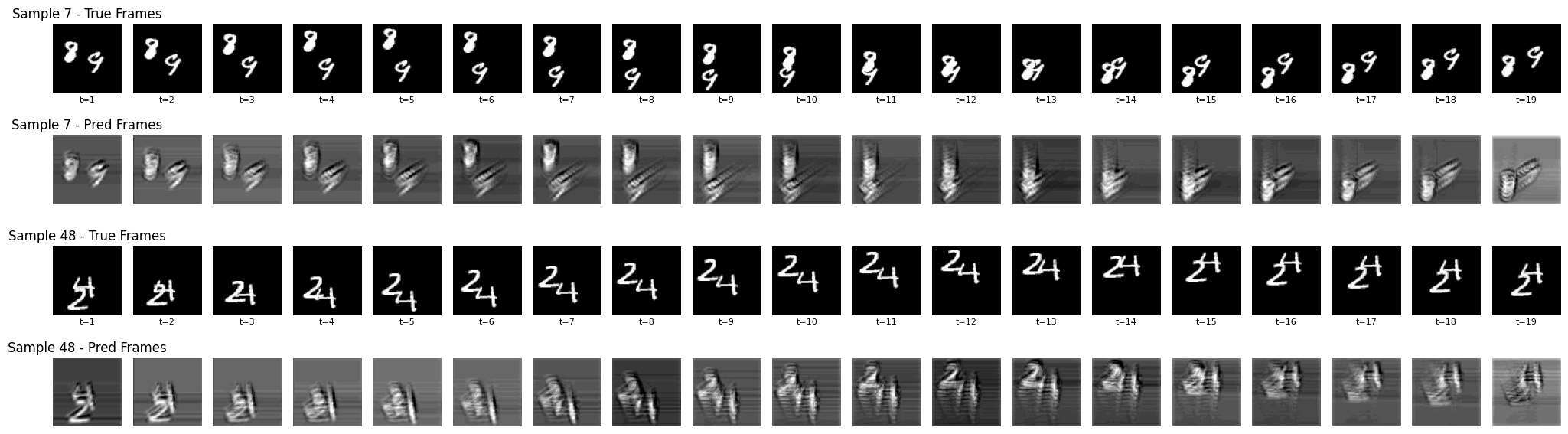}
    \label{fig:prediction-outcome}
\end{figure}

\section{Conclusion}

The proposed VAPAAD model represents a significant advancement in two-dimensional sequential prediction, effectively learning from past frames to forecast future sequences with high accuracy. Our experimental results substantiate the model's robust performance, particularly when integrating attention mechanisms, which outperform traditional training paths. However, the observed warping phenomenon in the predictions opens up new avenues for research, suggesting deeper investigation into how models perceive and interpret dynamic changes. The potential for adapting this mechanism to 3D applications is vast, ranging from film-to-film predictions to critical medical scenarios like tracking mutations in cancerous cells, or even predicting the flight trajectory of intercontinental ballistic missiles, highlighting the broad applicability and profound impact of our findings.

\section*{Data and Model Availability}

The Moving Dataset is publicly available \href{https://www.cs.toronto.edu/~nitish/unsupervised_video/}{here}. We are grateful for their work \cite{srivastava2015unsupervised}.

\section*{Funding}
No funding information available.

\section*{Authors' contributions}
Yiqiao Yin wrote the main manuscript text. Yiqiao Yin designed the experiment and ran the code. Yiqiao Yin collected the data and was responsible for the data processing pipeline. Yiqiao Yin contributed to the major design of the app backed by the architecture proposed in the paper. The author reviewed the manuscript. The author read and approved the final manuscript.

\section*{Acknowledgments}
The work is dedicated to Frances June Nunn, a life-long partner and a technical editor who also contributed to this paper.

\section*{Ethical Declarations}
The author declare no competing interests.

\bibliographystyle{unsrt}  
\bibliography{references}

\end{document}